\documentclass{article}
\usepackage{ijcai21}

\usepackage{times}
\usepackage{soul}
\usepackage{url}
\usepackage[hidelinks]{hyperref}
\usepackage[utf8]{inputenc}
\usepackage[small]{caption}

\usepackage{graphicx}
\graphicspath{{figures/}}

\usepackage{amsmath}
\usepackage{amsthm}
\usepackage{booktabs}
\usepackage{algorithm}
\usepackage{algorithmic}
\usepackage{multirow}
\usepackage{latexsym}
\usepackage{amscd}
\usepackage{amssymb}
\usepackage{mathrsfs}

\usepackage{cleveref}

\crefname{section}{Section}{Sections}
\crefname{subsection}{Subsection}{Subsections}
\crefname{algorithm}{Algorithm}{Algorithms}
\crefname{table}{Table}{Tables}
\crefname{figure}{Figure}{Figures}

\usepackage{autonum}

\usepackage{xcolor}

\newcommand{\R}{\mathbb{R}}
\newcommand{\N}{\mathbb{N}}

\newcommand{\XX}{\mathcal{X}}
\newcommand{\YY}{\mathcal{Y}}

\newcommand{\NN}{\mathcal{N}}
\newcommand{\MM}{\mathcal{M}}

\newcommand{\FF}{\mathcal{F}}

\renewcommand{\AA}{\mathcal{A}}

\newcommand{\TT}{\mathcal{T}}

\renewcommand{\epsilon}{\varepsilon}

\DeclareMathOperator*{\argmax}{arg\,max}

\newcommand{\theTitle}{Topological Uncertainty: Monitoring trained neural networks \\ through persistence of activation graphs}

\title{\theTitle}

\author{Content areas: Machine Learning:Deep Learning, Uncertainty in AI:Uncertainty Representations}

\author{
Théo Lacombe$^1$\footnote{Contact Author}\and
Yuichi Ike$^{2}$\and
Mathieu Carrière$^1$\and
Frédéric Chazal$^1$\and
Marc Glisse$^1$ \and 
Yuhei Umeda$^{2}$ \\
\affiliations
$^1$DataShape - Inria\\
$^2$Fujitsu Laboratories Ltd. \\
\emails
$^1$\{firstname.lastname\}@inria.com,
$^2$\{lastname.firstname\}@fujitsu.com
}

\begin{document}

\maketitle

\begin{abstract}
Although neural networks are capable of reaching astonishing performances on a wide variety of contexts, properly training networks on complicated tasks requires expertise and can be expensive from a computational perspective. 
In industrial applications, data coming from an open-world setting might widely differ from the benchmark datasets on which a network was trained. 
Being able to monitor the presence of such variations without retraining the network is of crucial importance. 
In this article, we develop a method to monitor trained neural networks based on the topological properties of their activation graphs. 
To each new observation, we assign a \emph{Topological Uncertainty}, a score that aims to assess the reliability of the predictions by investigating the whole network instead of its final layer only, as typically done by practitioners. 
Our approach entirely works at a post-training level and does not require any assumption on the network architecture, optimization scheme, nor the use of data augmentation or auxiliary datasets; and can be faithfully applied on a large range of network architectures and data types. 
We showcase experimentally the potential of Topological Uncertainty in the context of trained network selection, Out-Of-Distribution detection, and shift-detection, both on synthetic and real datasets of images and graphs.
\end{abstract}

\section{Introduction}

Over the last decade, Deep Learning (DL) has become the most popular approach to tackle complex machine learning tasks, opening the door to a broad range of industrial applications. 
Despite its undeniable strengths, monitoring the behavior of deep Neural Networks (NN) in real-life applications can be challenging. 
The more complex the architectures become, the stronger the predictive strengths of the networks, but the looser our grasp on their behaviors and weaknesses. 

Training a neural network requires task-specific expertise, is time consuming, and requires the use of high-end expensive hardware.
With the rise of companies providing model marketplaces (\textit{e.g.}, \cite[\S, Table 1]{kumar2020marketplace} or \href{https://www.tensorflow.org/hub}{\texttt{tensorflow-hub}}), it is now common that users only have access to fixed trained neural networks, with few information on the training process, and would rather avoid training the networks again. 

\begin{figure}[t]
    \centering
    \includegraphics[width=0.23\textwidth]{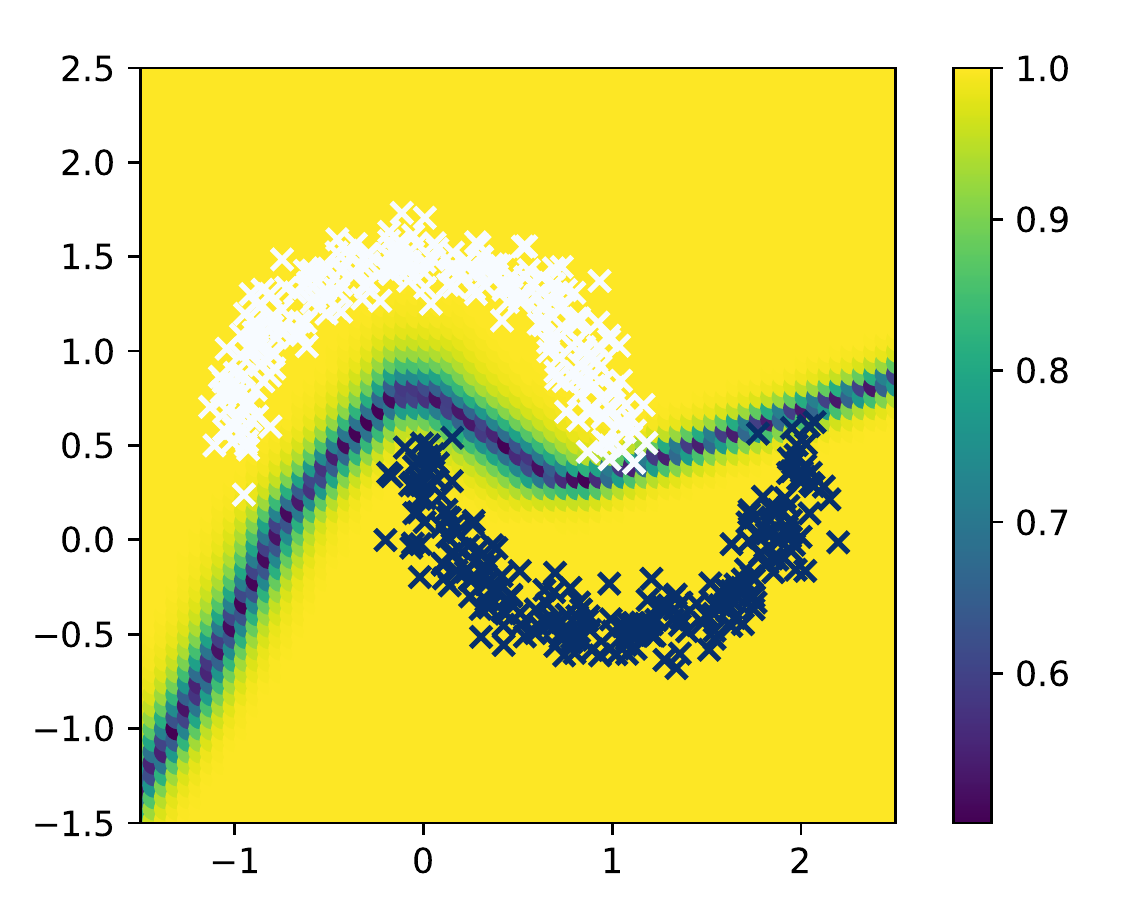}
    \includegraphics[width=0.23\textwidth]{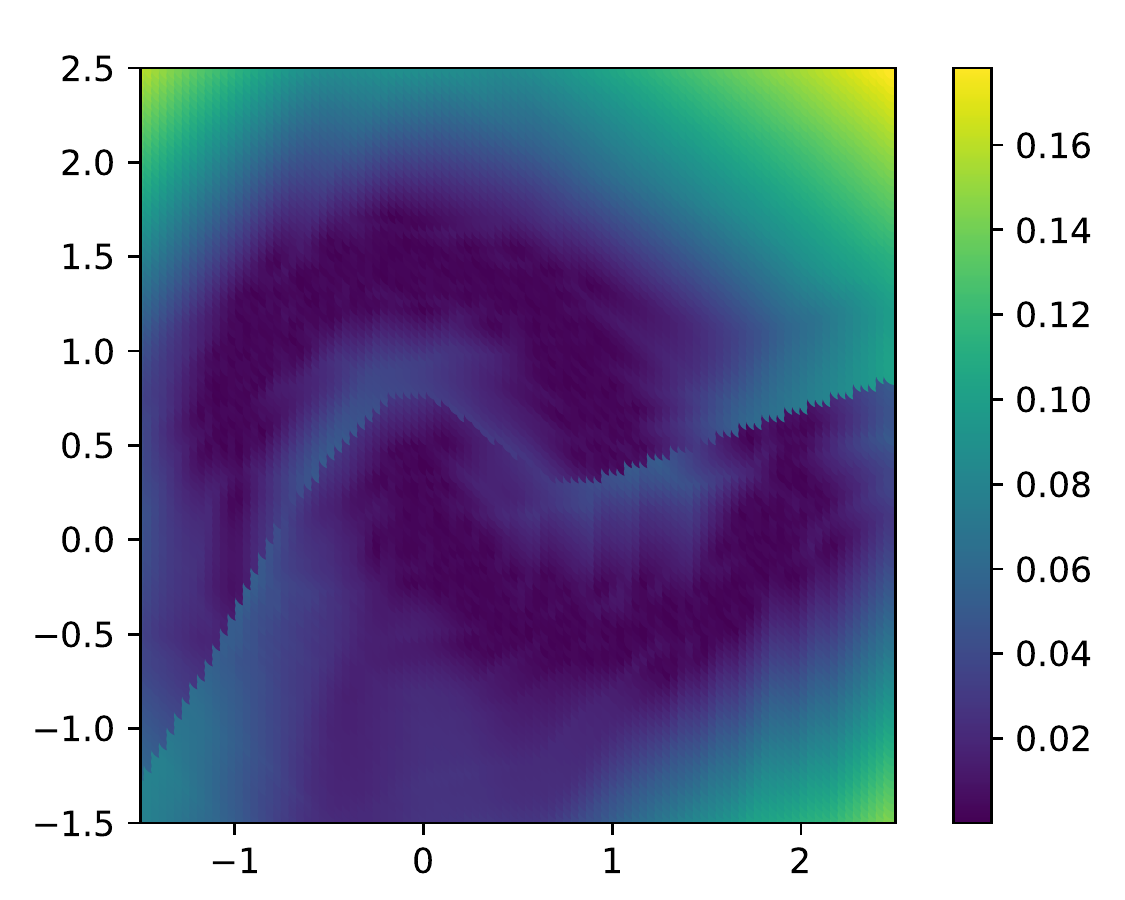}
    \caption{\textit{(Left)} Confidence (maximum in the final soft-max layer) of the predictions made by a neural network trained on the two-moons dataset (white and black crosses) over the plane. Away from the thin classification boundary (confidence value close to $0.5$), the network tends to produce \emph{over-confident predictions} (value close to $1$), even for points that are far away from the training data. \textit{(Right)} Topological distance between all activation graphs and the activation graphs computed on the train set (higher means less confident).}
    \label{fig:confidence_vs_topo_uncertainty}
\end{figure}

%We focus in this work on \emph{classification} problems, where the goal is to assign to a given \emph{observation} $x \in \XX$ a \emph{label} in a finite set $\YY = \{ 1 \dots K\}$. Given an observation $x \in \XX$, a neural networks (NN) $F$ outputs a probability distribution $F(x) = (p_1 \dots p_K)$ supported on $\YY$, where $p_i$ represents the confidence the network has in assigning the label $i \in \YY$ to $x$; making $i^* = \argmax\{F(x)\}$ the actual prediction made by the network and $p_{i^*}$ the confidence in the NN in its prediction. 

In practice, a network can perform well on a given learning problem---in the sense that it can achieve high accuracy on the training and test sets---, but lack reliability when used in real-life applications thereafter. 
One can think of, among other examples, \emph{adversarial attacks}, that are misinterpreted by neural networks with high confidence levels \cite{nguyen2015deep,akhtar2018threat,biggio2018wild}, \emph{calibration} issues leading to under- or over-confident predictions \cite{guo2017calibration,ovadia2019can,hein2019relu}, lack of \emph{robustness} to corruption or perturbation of the input data \cite{hendrycks2019benchmarking}.
% Although various techniques have been proposed to mitigate these issues in specific cases, they generally come with limited theoretical guarantees. 

% In this work, we consider that a user has access to a trained-network which is assumed to reach high accuracy on train and test sets. 
% The user wants to deploy the network in some real-life application where the distribution of new observations may differ from the training sets. 
% Following the phrasing of \cite{meinke2019towards}, we want to ``know when a network does not know'', that is assessing the reliability of the network predictions when it faces new observations, possibly different from the training dataset.\
% Although it might be tempting to use the network confidence as a way to monitor it and to detect peculiar observations (\textit{e.g.}~adversaries, distribution shift, Out-Of-Distribution (OOD) inputs), this approach is known to unfortunately fail \cite{hein2019relu} without specific tweaking of the network during its training phase. 

% Our goal is specifically to avoid relying on such tweaking, and to show that one can still obtain relevant information from looking at the whole network instead of restricting to its last layer.

% Given a well-trained neural network, The exposition of NN to observations that were not part of the training (or validation) sets has known a gain of interest in the very last years \cite{hsu2020generalized,chen2020robust}.  

\paragraph{Related work.}
Many techniques have been proposed to improve or monitor the behavior of NN deployed in real-life applications. 
Most require specific actions taken during the training phase of the network; for instance via \emph{data augmentation} \cite{shorten2019survey}), the use of large auxiliary datasets\footnote{In particular, the \texttt{80 million tiny images dataset} \cite{torralba200880}, used as an auxiliary dataset in state-of-the-art OOD detection techniques \cite{chen2020robust}, has been withdrawn for ethical concerns. This situation illustrates unexpected limitations when relying on such datasets to calibrate neural networks.} \cite{hendrycks2019oe}, modifications of the network architecture \cite{devries2018learning} or its objective function \cite{atzmon2019controlling,joost2020uncertainty}, or using several networks at once to produce predictions \cite{lakshminarayanan2017simple}. %,havasi2020training 
The approach we introduce in this article focuses on acting at a post-training level only, and since our goal is to benchmark it against the use of confidence alone in a similar setting, we use, in our experimental results, the baseline introduced in \cite{hendrycks2017baseline}, which proposes to monitor NNs based on their confidences and on the idea that low-confidence predictions may account for anomalies.

Using topological quantities to investigate NN properties has experienced a growth of interest recently (see for instance \cite{guss2018characterizing,carlsson2020topological}). 
In \cite{gebhart2017adversary,gebhart2019characterizing}, the authors introduce the notion of activation graphs and showcase their use in the context of adversarial attacks. 
We mix this idea with \cite{rieck2019neural}, which proposes to investigate topological properties of NN layer-wise\footnote{A more detailed comparison between our work and \cite{rieck2019neural} is deferred to the appendix}. 
A topological score based on an estimation of the NN decision boundary has been proposed in \cite{ramamurthy2019topological} to perform trained network selection, an idea we adapt in \cref{subsec:modelselec}.

\paragraph{Contributions.}

%Our goal is to propose a method that is independent from the way the network has been trained and thus make no assumption on potential robustness properties of the network. 
In this work, we propose a new approach to monitor trained NNs by leveraging the topological properties of \emph{activation graphs}. 
%% Shorten a bit the following %%
Our main goal is to showcase the potential benefits of investigating network predictions through the lens of the whole network instead of looking at its \emph{confidence} encoded by its final layer only as usually done in practice.

To that aim, we introduce \emph{Topological Uncertainty} (TU), a simple topological quantity that, for a given NN and a new observation, encodes how the network ``reacts'' to this observation, and whether this reaction is similar to the one on training data. 
Our approach does not require any assumption on the network training phase nor the type of input data, and can thus be deployed in a wide variety of settings.
Furthermore, it only relies on computating maximum spanning trees (MST), leading to a simple and efficient implementation.
%%% Can we improve the above sentence a bit? %%%

Experimentally, we show that TU can be used to monitor trained NNs and detect Out-of-Distribution (OOD) or shifted samples when deployed in real-life applications. 
Our results suggest that TU can drastically improve on a standard baseline based on the network confidence in different situations.
Our implementation will be made publicly available.

\section{Background}
\label{sec:background}

%In this section, we present a short background that intends to serve some basics for this article. 

\subsection{Neural networks}

For the sake of clarity, we restrict our presentation to \emph{sequential} neural networks, although our approach (detailed in \cref{sec:methodology}) can be generalized to more general architectures, \textit{e.g.}, recurrent neural networks. 
We also restrict to classification tasks; let $d$ denote the dimension of the input space and $K$ be the number of classes. 
A (sequential) neural network (NN) learns a function $F \colon \R^d \to \R^K$ that can be written as $F = f_L \circ \dots \circ f_1$, where the $(f_\ell)_{\ell=1}^L$ are elementary blocks defined for $\ell =1, \dots, L-1$ as
\begin{align}
    x_{\ell+1} = f_\ell(x_\ell) = \sigma_\ell(W_\ell \cdot x_\ell + b_\ell),
\end{align}
with $W_\ell \in \R^{d_\ell \times d_{\ell+1}}$, $b_\ell \in \R^{d_{\ell+1}}$, and $(\sigma_\ell)_\ell$ are activation maps\footnote{Note that this formalism encompasses both fully-connected and convolutional layers that are routinely used by practitioners.}, \textit{e.g.}, $\sigma_\ell = \mathrm{ReLU} \colon x \mapsto \max\{x,0\}$.
% We write $x=x_1 \in \R^{d_1}=\R^d$ for an input instance or observation. 
% as particular instances. 
In classification tasks, the final activation $f_L = \sigma_L$ is usually taken to be the soft-max function, so that the output $x_L = F(x)$ can be understood as a probability distribution on $\{1, \dots, K\}$ whose entries $(F(x)_k)_k$ indicate the likelihood that $x$ belongs to class $k$. 
The predicted class is thus $\argmax_k\{ F(x)_k \}$, while the \emph{confidence} that the network has in its prediction is $\max_k\{ F(x)_k \} \in [0,1]$. 
Given a \emph{training set} of observations and labels $(X^\text{train}, Y^\text{train}) = (x_i,y_i)_{i=1}^{N_\mathrm{train}}$ distributed according to some (unknown) joint law $(\XX, \YY)$, the network parameters $W_\ell, b_\ell$ are optimized to minimize the loss $\sum \mathcal{L} (F(x_i), y_i)$ for some loss function $\mathcal{L}$ (\textit{e.g.}, the categorical cross-entropy). 
The (training) accuracy of $F$ is defined as $\frac{1}{N_\mathrm{train}} \sum_{i=1}^{N_\mathrm{train}} 1_{\argmax(F(x_i)) = y_i}$. 
% Note that accuracy and loss are not equal, though correlated. 

\subsection{Activation graphs and topological descriptors}

\paragraph{Activation graphs.} 

Let us consider a neural network $F$ and two layers of size $d_\ell$ and $d_{\ell+1}$ respectively, connected by a matrix $W_\ell \in \R^{d_\ell \times d_{\ell+1}}$. 
One can build a bipartite graph $G_\ell$ whose vertex set is $V_\ell \sqcup V_{\ell+1}$ with $|V_\ell|=d_\ell$ and $|V_{\ell+1}| = d_{\ell+1}$, and edge set is $E_\ell = V_\ell \times V_{\ell+1}$. 
Following \cite{gebhart2019characterizing}, given an instance $x \in \R^d$, one can associate to each edge $(i,j) \in E_\ell$ the weight $|W_\ell(i,j) \cdot x_\ell(i)|$, where $x_\ell(i)$ (resp.\ $W_\ell(i,j)$) denotes the $i$-th coordinate of $x_\ell \in \R^{d_\ell}$ (resp.\ entry $(i,j)$ of $W_\ell \in \R^{d_\ell \times d_{\ell+1}}$). 
Intuitively, the quantity $|W_\ell(i,j) \cdot x_\ell(i)|$ encodes how much the observation $x$ ``activates'' the connection between the $i$-th unit of the $\ell$-th layer and the $j$-th unit of the $(\ell+1)$-th layer of $F$. 
In this way, we obtain a sequence of bipartite graphs $(G_\ell(x,F))_\ell$ called the \emph{activation graphs} of the pair $(x,F)$, whose vertices are $V_\ell \sqcup V_{\ell +1}$ and edges weights are given by the aforementioned formula.

\paragraph{Maximum spanning trees and persistence diagrams.}

To summarize the information contained in these possibly large graphs in a quantitative way, we rely on topological descriptors called \emph{persistence diagrams}, coming from the Topological Data Analysis (TDA) literature. 
A formal introduction to TDA is not required in this work (we refer to the appendix for a more general introduction): in our specific case, persistence diagrams can be directly defined as the distribution of weights of a \emph{maximum spanning tree} (MST). 
We recall that given a connected graph $G$ with $N+1$ vertices, a MST is a connected acyclic sub-graph of $G$ sharing the same set of vertices such that the sum of the $N$ edge weights is maximal among such sub-graphs. MST can be computed efficiently, namely in quasilinear time with respect to the number of edges in $G$. 
Given the ordered weights $w_1 \geq \dots \geq w_{N}$ of a MST built on top of a graph $G$, its persistence diagram is the one-dimensional probability distribution
\begin{align}
    \mu(G) := \frac{1}{N} \sum_{i=1}^{N} \delta_{w_i},
\end{align}
where $\delta_{w_i}$ denotes the Dirac mass at $w_i \in \R$. See \cref{fig:pipeline} for an illustration.

\paragraph{Comparing and averaging persistence diagrams.} 

The standard way to compare persistence diagrams relies on \emph{optimal partial matching metrics}. 
The choice of such metrics is motivated by stability theorems %\cite{cohen2007stability,chazal2016structure} 
that, in our context, imply that the map $x \mapsto \mu(G_\ell(x,F))$ is Lipschitz with Lipschitz constant that only depends on the network architecture and weights\footnote{\label{note}We refer to the appendix for details and proofs.} (and not on properties of the distribution of $x \sim \XX$ for instance). 
The computation of these metrics is in general challenging \cite{kerber2017geometry}. 
However, in our specific setting, the distance $\mathrm{Dist}(\mu,\nu)$ between two diagrams $\mu = \frac{1}{N} \sum_{i=1}^N \delta_{w_i}$ and $\nu = \frac{1}{N} \sum_{j=1}^N \delta_{w'_j}$ can be simply obtained by computing a 1D-optimal matching\textsuperscript{\ref{note}}, %\footnote{See footnote \ref{note}.}, 
which in turn only requires to match points in increasing order, leading to the simple formula 
\begin{align}
    \mathrm{Dist}(\mu,\nu)^2 = \frac{1}{N} \sum_{i=1}^N |w_i - w'_i|^2,
\end{align}
where $w_1 \geq w_2 \geq \dots \geq w_N$ and $w'_1 \geq w'_2 \geq \dots \geq w'_N$. 
With this metric comes a notion of \emph{average persistence diagram}, called a \emph{Fr\'echet mean} \cite{turner2014frechet}: a Fr\'echet mean $\overline{\mu}$ of a set of $M$ diagrams $\mu_1, \dots, \mu_M$ is a minimizer of $\nu \mapsto \sum_{m=1}^M \mathrm{Dist}(\mu_m,\nu)^2$, which in our context simply reads
\begin{align}
\overline{\mu} := \frac{1}{N} \sum_{i=1}^N \delta_{\bar w_i}, %\overline{w_i}},
\label{eq:FrechetMean}
\end{align}
where $\bar w_i %\overline{w_i} 
= \frac{1}{M} \sum_{m=1}^M w_i^{(m)}$ and $w_i^{(m)}$ denotes the $i$-th point of $\mu_m$. 
The Fr\'echet mean provides a geometric way to concisely summarize the information contained in a set of persistence diagrams. %the topology of the activated graphs obtained on our trained network.

%%%%%%%%%%%%%%%%%%%%%%%
\section{Topological Uncertainty (TU)}
\label{sec:methodology}

\begin{figure*}[t]
    \centering
    \includegraphics[width=0.9\textwidth]{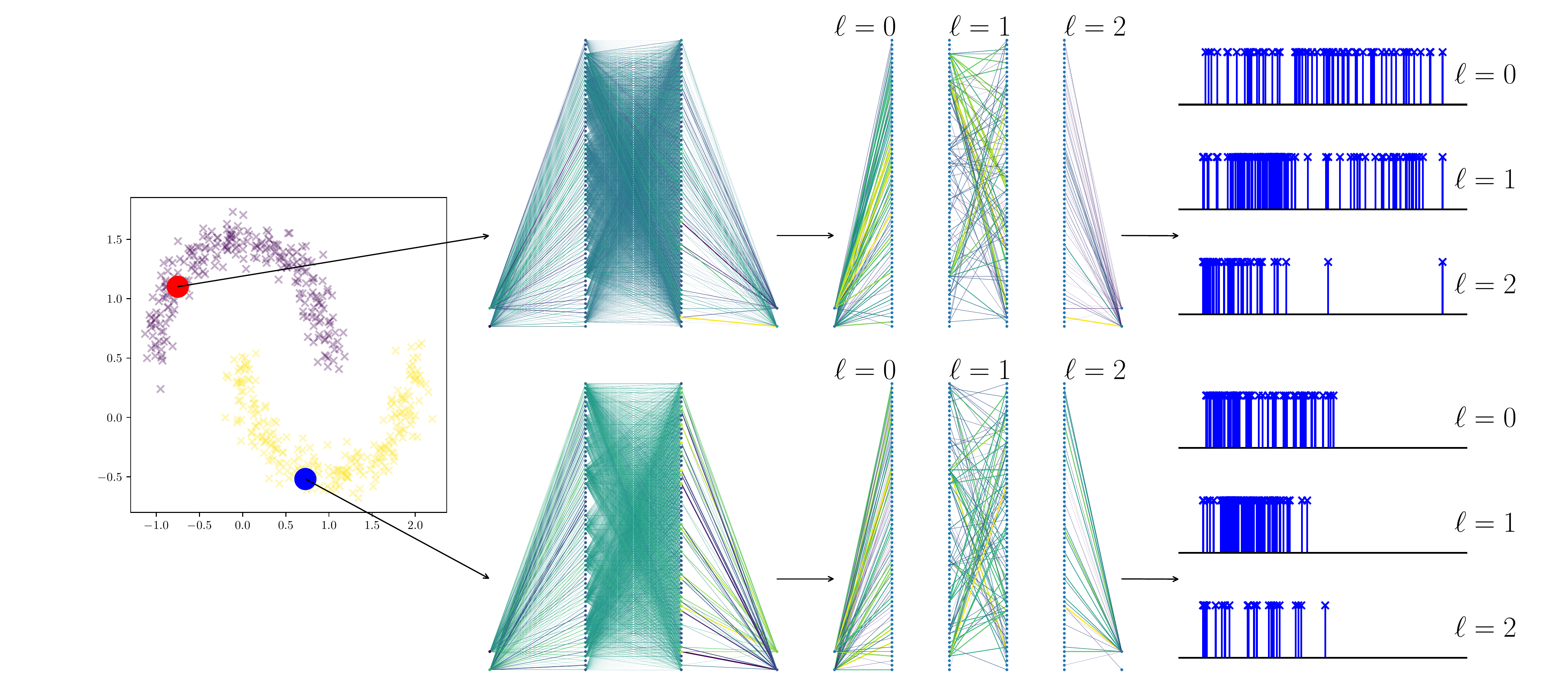}
    \vspace{-0.3cm}
    \caption{Pipeline presented in this article: Each observation activates the network with a weight $|W_\ell(i,j) \cdot x_\ell(i)|$ on the edge connecting the $i$-th unit in the $\ell$-th layer to the $j$-th unit of the $(\ell+1)$-th layer. A maximum spanning tree is then computed for each layer $\ell$ of the network, whose distribution of weights provides a corresponding \emph{persistence diagram}. On this example, the network used is a simple network with two hidden-layers of 64 units each with ReLU activation, each layer is fully-connected (dense matrix).}
    \label{fig:pipeline}
\end{figure*}

Building on the material introduced in \cref{sec:background}, we propose the following pipeline, which is summarized in \cref{fig:pipeline}. 
Given a trained network $F$ and an observation $x$, we build a sequence of activation graphs $G_1(x,F), \dots, G_{L-1}(x,F)$. 
We then compute a MST of each $G_\ell(x,F)$, which in turn induces a persistence diagram $D_\ell(x,F) := \mu(G_\ell(x,F))$.

\paragraph{Topological Uncertainty.} 

Given a network $F \colon \R^d \to \R^K$ trained on a set $X^{\text{train}}$, one can store the corresponding sequence of diagrams $(D_\ell(x^\text{train}, F))_\ell$ for each $x^\text{train} \in X^\text{train}$. 
These diagrams summarize how the network is activated by the training data. 
Thus, given a new observation $x$ with $\argmax(F(x)) = k$, one can compute the sequence $(D_\ell(x,F))_\ell$ and then the quantity
% \begin{align}
%      \min \{ \mathrm{Dist}(D_\ell(x,F), D_\ell(x^\text{train},F)) \;:\; x^\text{train} \in X^\text{train}, \argmax(F(x^\text{train})) = k \},
% \end{align}
\begin{align}
     \min_{\substack{x^\text{train} \in X^\text{train}, \\ \argmax( F(x^\text{train})) = k}} \mathrm{Dist} \left(D_\ell(x,F), D_\ell(x^\text{train},F) \right),
\end{align}
% \begin{equation}
%   \DD_{\ell,k}^\text{train} := \{ D_\ell(x^\text{train}, F),\ \argmax(F(x^\text{train})) = k \},
% \end{equation}
that is, comparing $D_\ell(x,F)$ to the diagrams of training observations that share the same predicted label than $x$. 
\cref{fig:confidence_vs_topo_uncertainty} (right) shows how this quantity evolves over the plane, and how, contrary to the network confidence (left), it allows one to detect instances that are far away from the training distribution. 

Since storing the whole set of training diagrams for each class and each layer $\{ D_\ell(x^\text{train}, F) \;:\; \argmax(F(x^\text{train})) = k \}$ might be inefficient in practice, we propose to summarize these sets through their respective Fr\'echet means $\overline{D^{\text{train}}_{\ell,k}}$. 
For a new observation $x \in \R^d$, let $k(x) = \argmax(F(x))$ be its predicted class, and $(D_\ell(x,F))_\ell$ the corresponding persistence diagrams. The \emph{Topological Uncertainty} of $x$ is defined to be
\begin{equation}
    \mathrm{TU}(x, F) := \frac{1}{L} \sum_{\ell=1}^L \mathrm{Dist} \left( D_{\ell}(x,F) , \overline{D^{\text{train}}_{\ell,k(x)}} \right),
    \label{eq:TU}
\end{equation}
which is the average distance over layers between the persistence diagrams of the activation graphs of $(x,F)$ and the average diagrams stored from the training set. 
Having a low TU suggests that $x$ activates the network $F$ in a similar way to the points in $X^\text{train}$ whose class predicted by $F$ was the same as $x$. 
Conversely, an observation with a high TU (although being possibly classified with a high confidence) is likely to account for an OOD sample as it activates the network in an unusual manner.

\paragraph{Remarks.} 

Our definition of activation graphs differs from the one introduced in \cite{gebhart2019characterizing}, as we build one activation graph for each layer, instead of a single, possibly very large, graph on the whole network. Note also that the definition of TU can be declined in a variety of ways. 
First, one does not necessarily need to work with all layers $1 \leq \ell \leq L$, but can only consider a subset of those. 
Similarly, one can estimate Fr\'echet means $\overline{D^\text{train}_{\ell,k}}$ using only a subset of the training data. 
These techniques might be of interest when dealing with large datasets and deep networks. 
One could also replace the Fr\'echet mean $\overline{D^\text{train}_{\ell,k}}$ by some other diagram of interest; in particular, using the empty diagram instead allows us to retrieve a quantity analog to the \emph{Neural persistence} introduced in \cite{rieck2019neural}. 
On an other note, there are other methods to build persistence diagrams on top of activation graphs that may lead to richer topological descriptors, but our construction has the advantage of returning diagrams supported on the real line (instead of the plane as it usually occurs in TDA) with fixed number of points, which dramatically simplifies the computation of distances and Fr\'echet means, and makes the process efficient practically.

\section{Experiments}
\label{sec:expe}

This section showcases the use of TU in different contexts: trained network selection (\S\ref{subsec:modelselec}), monitoring of trained networks that achieve large train and test accuracies but have not been tweaked to be robust to Out-Of-Distribution observations (\S\ref{subsec:ood}) or distribution shift (\S\ref{subsec:shift}).
%We show how TU can drastically improve on the network confidence $\max(F(x))$ as a way to detect such samples.

\paragraph{Datasets and experimental setting.} 

We use standard, publicly available, datasets of graphs and images. 
\texttt{MNIST}, \texttt{Fashion-MNIST}, \texttt{CIFAR-10}, \texttt{SVHN}, \texttt{DTD} are datasets of images, while \texttt{MUTAG} and \texttt{COX2} are datasets of graphs coming from a chemical framework. We also build two OOD image datasets, \texttt{Gaussian} and \texttt{Uniform}, by randomly sampling pixel values following a Gaussian distribution (resp.~uniform on the unit cube).
A detailed report of datasets and experimental settings (data preprocessing, network architectures and training parameters, etc.) can be found in the appendix. 

\subsection{Trained network selection for unlabeled data}
\label{subsec:modelselec}

\begin{figure}[t]
    \centering
    \includegraphics[width=0.3\textwidth]{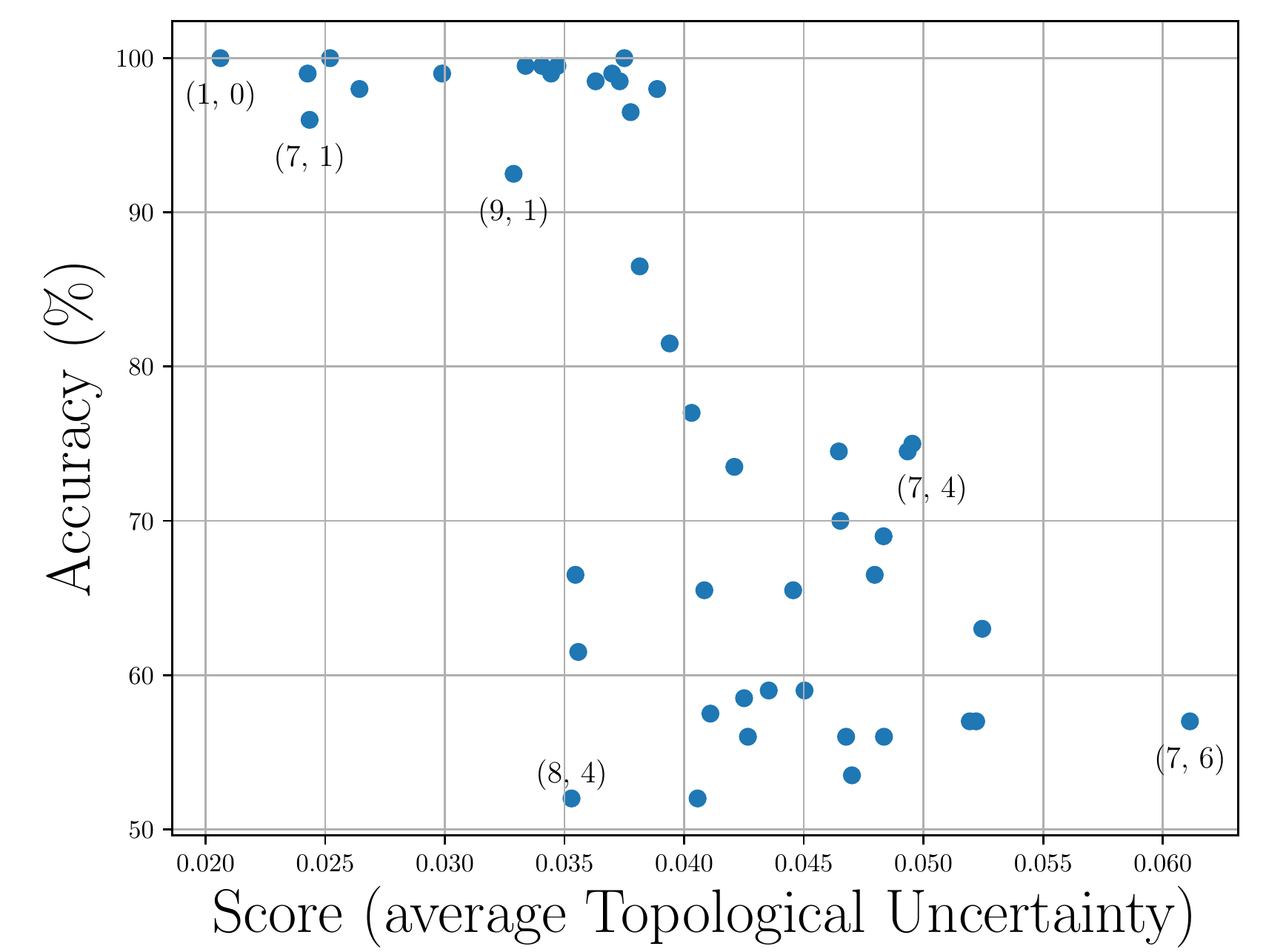}
    \vspace{-0.3cm}
    \caption{Score and accuracies obtained on 45 models trained on the MNIST dataset when fed with a set $X_{1,0}$ of images representing $0$ and $1$ digits. The point annotations refer to the values $(i,j)$ on which the network was initially trained. For instance, a network trained on $7$ and $4$ has a score of $0.49$ and an accuracy of $\sim75$\% on $X_{1,0}$.}
    \label{fig:score_10}
\end{figure}

In this subsection, we showcase our method in the context of trained network selection through an experiment proposed in \cite[\S 4.3.2]{ramamurthy2019topological}. 
Given a dataset with 10 classes (here, \texttt{MNIST} or \texttt{Fashion-MNIST}), we train $45$ NNs on the binary classification problems $i$ \textit{vs.}~$j$ for each pair of classes $(i,j)$ with $i > j$, and store the average persistence diagrams of the activation graphs for the different layers and classes as explained in \cref{sec:methodology}. 
These networks are denoted by $F_{ij}$ in the following, and consistently reach high accuracies on their respective training and test sets given the simplicity of the considered tasks. 
Then, for each pair of classes $k_1 > k_2$, we sample a set of new observations $X_{k_1, k_2}$ made of $200$ instances sampled from the \emph{test} set of the initial dataset (in particular, these observations have not been seen during the training of any of the $(F_{ij})_{ij}$) whose labels are $k_1$ and $k_2$. 
Assume now that $k_1, k_2$ and the labels of our new observations are unknown. 
The goal is to select a network that is likely to perform well on $X_{k_1,k_2}$ among the $(F_{ij})_{ij}$. 
To that aim, we compute a \emph{score} for each pairing $(k_1, k_2) \leftrightarrow (i,j)$, which is defined as the average TU (see Eq.~\eqref{eq:TU}) when feeding $F_{ij}$ with $X_{k_1, k_2}$. 
A low score between $(i,j)$ and $(k_1, k_2)$ suggests that $X_{k_1, k_2}$ activates $F_{ij}$ in a ``known'' way, thus that $F_{ij}$ is likely to be a relevant classifier for $X_{k_1, k_2}$, while a high score suggests that the data in $X_{k_1, k_2}$ is likely to be different from the one on which $F_{i,j}$ was trained, making it a less relevant choice. 
\cref{fig:score_10} plots the couple (scores, accuracies) obtained on \texttt{MNIST} when taking $(k_1, k_2) = (1, 0)$, that is, we feed networks that have been trained to classify between $i$ and $j$ handwritten digits with images representing $0$ and $1$ digits. 
Not surprisingly, $F_{1,0}$ achieves both a small TU (thus would be selected by our method) and a high accuracy. 
On the other hand, using the model $F_{7,6}$ on $X_{1,0}$ leads to the highest score and a low accuracy. 

To quantitatively evaluate the benefit of using our score to select a model, we use the the metric proposed by \cite{ramamurthy2019topological}: the difference between the mean accuracy obtained using the 5 models that have the lowest scores and the 5 models that have the highest scores. 
We obtain a $+14.8$\%-increase on \texttt{MNIST} and a $+12.6$\%-increase on \texttt{Fashion-MNIST}. 
These positive values indicate that, on average, using low TU as a criterion helps to select a better model to classify our new set of observations. As a comparison, using the average network confidence as a score leads to $+6.6\%$ on \texttt{MNIST} and $+4.7\%$ on \texttt{Fashion-MNIST}, indicating that using (high) confidence to select a model would be less relevant than using TU, on average. See the appendix for a complementary discussion.

\paragraph{Remark.} 

Our setting differs from the one of \cite{ramamurthy2019topological}: in the latter work, the method requires to have access to true labels on the set of new observations $X_{k_1, k_2}$ (which we do not), but do not need to evaluate $F_{ij}(x),\ x \in X_{k_1, k_2}$ on each model $F_{ij}$ (which we do). 
To that respect, we stress the complementary aspect of both methods.

\subsection{Detection of Out-Of-Distribution samples}
\label{subsec:ood}
\begin{figure*}
    \centering
    \includegraphics[width=0.8\textwidth]{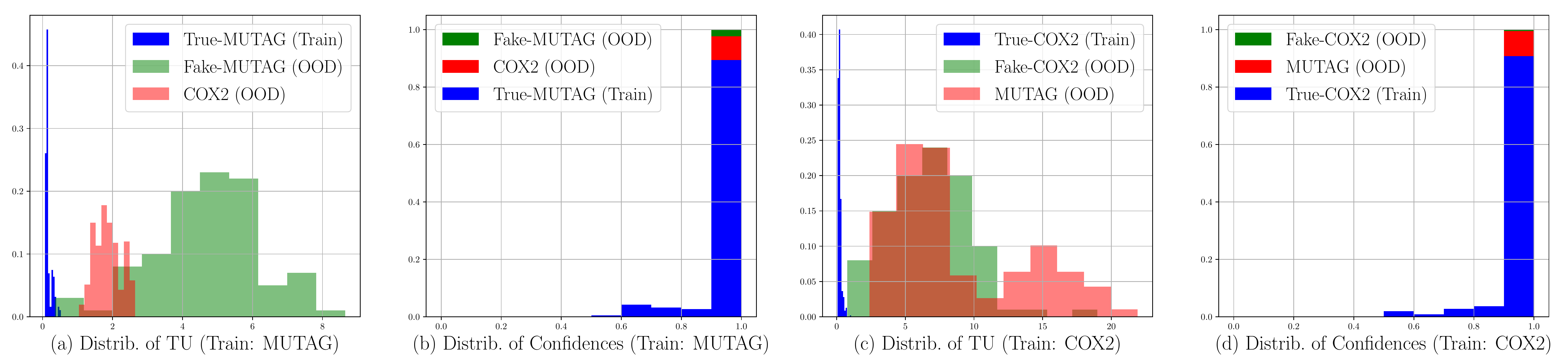}
    \vspace{-0.3cm}
    \caption{\textit{(a)} Distributions of Topological Uncertainties (TU) of a network trained on the \texttt{MUTAG} dataset. Blue distribution corresponds to $\TT^\text{train}$. Green and red distributions correspond to topological uncertainties of observations coming from \texttt{Fake-MUTAG} and \texttt{COX2} datasets respectively. \textit{(b)} Distributions of network confidences ($\max\{F(x)_k\}$). The network makes overconfident predictions, especially on OOD datasets that are classified almost systematically with a confidence of $1$. \textit{(c,d)} Similar plots for the \texttt{COX2} dataset. }
    \label{fig:OOD}
\end{figure*}

The following experiment illustrates the behavior of TU when a trained network faces Out-Of-Distribution (OOD) observations, that is, observations that are not distributed according to the training distribution. 
To demonstrate the flexibility of our approach, we present the experiment in the context of graph classification, relying on the \texttt{COX2} and the \texttt{MUTAG} graph datasets. Complementary results on image datasets can be found in \cref{tab:ood-graphs} and in the appendix.
Working with these datasets is motivated by the possibility of using simple networks while achieving reasonably high accuracies which are near state-of-the-art on these sets. 
To train and evaluate our networks, we extract $40$ spectral features from graphs---thus representing a graph by a vector in $\R^{40}$---following a procedure proposed in \cite{carriere2020perslay}. 
See the appendix for details.
%\footnote{We reach $88$\% on \texttt{MUTAG} and $78$\% on \texttt{COX2} respectively.}

For both datasets, we build a set of 100 \emph{fake graphs} in the following way. 
Let $\NN$ and $\MM$ be the distributions of number of vertices and number of edges respectively in a given dataset (\textit{e.g.}, graphs in \texttt{MUTAG} have on average $17.9 \pm 4.6$ vertices and $19.8 \pm 5.7$ edges). 
Fake graphs are sampled as Erd{\H o}s-Renyi graphs of parameters $(n, m/n^2)$, with $n \sim \NN$, $m \sim \MM$, thus by construction fake graphs have (on average) the same number of vertices and edges as graphs from the training dataset. 
These sets are referred to as \texttt{Fake-MUTAG} and \texttt{Fake-COX2}, respectively.

\begin{table}[t]
\resizebox{\columnwidth}{!}{%
\begin{tabular}{llcccc}
\hline
                                             &                                 & \multicolumn{2}{c}{Baseline} & \multicolumn{2}{c}{Topological Uncertainty} \\ \hline
\multicolumn{1}{c}{Training data}          & \multicolumn{1}{c}{OOD data}   & FPR (TPR 95\%)    $\downarrow$                        & AUC $\uparrow$               & FPR (TPR 95\%) $\downarrow$           & AUC $\uparrow$            \\ \hline
\multicolumn{1}{l}{\multirow{2}{*}{MUTAG}} & \multicolumn{1}{l}{Fake-MUTAG} & 98.4                                          &  1.7                              & {\bf 0.0}                          & {\bf 99.8}            \\ \cline{2-6} 
\multicolumn{1}{l}{}                       & \multicolumn{1}{l}{COX2}       & 93.0                                          & 31                               & {\bf 0.0}                          & {\bf 100.0}           \\ \hline
\multicolumn{1}{l}{\multirow{2}{*}{COX2}}  & \multicolumn{1}{l}{Fake-COX2}  & 91.2                                          &  1.4                              &           {\bf 0.0}                & {\bf 99.9}                \\ \cline{2-6} 
\multicolumn{1}{l}{}                       & \multicolumn{1}{l}{MUTAG}      & 91.2                                          & 1.1                               &       {\bf 0.0}                    & {\bf 100.0}                \\ \hline

\multicolumn{1}{l}{\multirow{6}{*}{CIFAR-10}} & \multicolumn{1}{l}{FMNIST}   & 93.6 & 54.9 & {\bf 65.6} & {\bf 86.4}  \\ \cline{2-6} 
\multicolumn{1}{l}{}                          & \multicolumn{1}{l}{MNIST}    & 94.7 & 58.3 & {\bf 25.4}  & {\bf 94.7} \\ \cline{2-6}
\multicolumn{1}{l}{}                          & \multicolumn{1}{l}{SVHN}     & 90.6 & 27.6 & {\bf 83.6}  & {\bf 65.8} \\ \cline{2-6}
\multicolumn{1}{l}{}                          & \multicolumn{1}{l}{DTD}      & 90.9 & 32.6 & 90.3        & {\bf 57.3} \\ \cline{2-6}
\multicolumn{1}{l}{}                          & \multicolumn{1}{l}{Uniform}  & 91.5 & 31.8 & {\bf 59.1}       & {\bf 80.2} \\ \cline{2-6}
\multicolumn{1}{l}{}                          & \multicolumn{1}{l}{Gaussian} & 91.0 & 27.2 & {\bf 18.8}  & {\bf 88.2} \\ \hline

\end{tabular}}

\vspace{-0.3cm}
\caption{Comparison between the baseline OOD-detector based on network confidence and our TU-based classifier on graph datasets (first two rows) and on image datasets for a network trained on \texttt{CIFAR-10} (third row). $\uparrow$: higher is better, $\downarrow$: lower is better.}
\label{tab:ood-graphs}
\end{table}

Now, given a network $F_\texttt{MUTAG}$ trained on \texttt{MUTAG} (resp.\ $F_\texttt{COX2}$ trained on  \texttt{COX2}), we store the average persistence diagrams of each classes. It allows
us to compute the corresponding distribution of TUs $\TT^\text{train} = \{ \mathrm{TU}(x, F_\texttt{MUTAG}) \;:\; x \in \texttt{MUTAG}\}$ (resp.\ $\mathrm{TU}(x,F_\texttt{COX2})$). Similarly, we evaluate the TUs of graphs from the \texttt{Fake-MUTAG} (resp.\ \texttt{Fake-COX2}) and from \texttt{COX2} (resp.\ \texttt{MUTAG}). 
These distributions are shown on \cref{fig:OOD} (second and fourth plots, respectively). 
As expected, the TUs of training inputs $\TT^\text{train}$ are concentrated around low values. 
Conversely, the TUs of OOD graphs (both from the Fake dataset and the second graph dataset) are significantly higher. 
Despite these important differences in terms of TU, the network still shows confidence near 100\% (first and third plots) over all the OOD datasets, making this quantity impractical for detecting OOD samples.

To quantify this intuition, we propose a simple OOD detector.
Let $F$ be a network trained on \texttt{MUTAG} or \texttt{COX2}, with distribution of training TUs $\TT^\text{train}$. 
A new observation $x$ is classified as an OOD sample if $\mathrm{TU}(x, F)$ is larger than the $q$-th quantile of $\TT^\text{train}$. 
This classifier can be evaluated with standard metrics used in OOD detection experiments: the \emph{False Positive Rate at 95\% of True Positive Rate} (FPR at 95\%TPR), and the Area Under the ROC Curve (AUC). 
We compare our approach with the baseline introduced in \cite{hendrycks2017baseline} based on confidence only: a point is classified as an OOD sample if its confidence is \emph{lower} than the $q$-th quantile of the distribution of confidences of training samples. 
As recorded in \cref{tab:ood-graphs}, this baseline performs poorly on these graph datasets, which is explained by the fact that (perhaps surprisingly) the assumption of \textit{loc.~cit.}\ that training samples tend to be assigned a larger confidence than OOD-ones is not satisfied in this experiment. 
In the third row of this Table, we provide similar results for a network trained on \texttt{CIFAR-10} using other image datasets as OOD sets. 
Although in this setting TU is not as efficient as it is on graph datasets, it still improves on the baseline reliably. 
%See the supplementary material for details.
%Results have been averaged over 100 runs to illustrate the robustness of our approach with respect to the network training. 
%We provide in the supplementary material a similar experiments involving image datasets. 

\subsection{Sensitivity to shifts in sample distribution}
\label{subsec:shift}
\begin{figure*}[th]
    \centering    
    \includegraphics[width=0.9\textwidth]{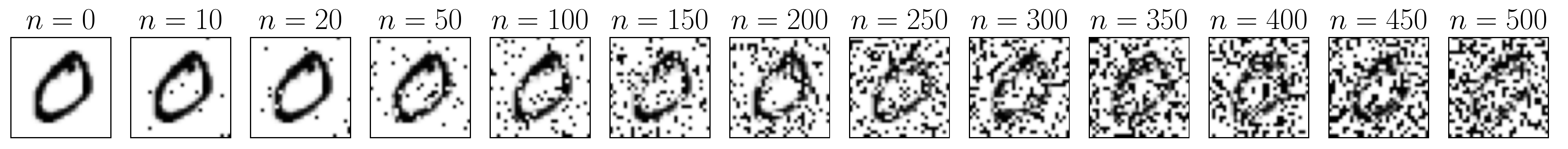}
    \includegraphics[width=0.8\textwidth]{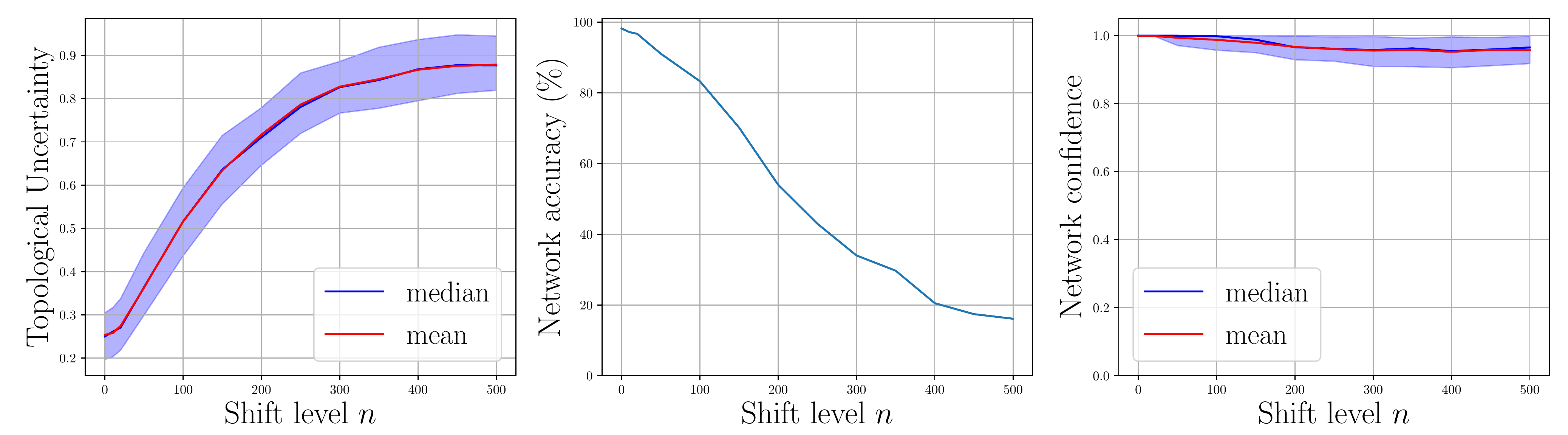}
    \vspace{-0.3cm}
    \caption{\textit{(Top row)} A $0$ digit from the \texttt{MNIST} dataset exposed to increasing level of shift (pixel corruption). \textit{(Bottom row)}, \textit{(Left)}. The TU (with $0.1$ and $0.9$ quantiles) of corrupted inputs in the \texttt{MNIST} dataset with respect to the corruption level $n$. \textit{(middle)} The accuracy of the network on these data (that is, the proportion of observations that are still correctly classified). \textit{(right)} The confidence of the network in its predictions. Although the accuracy is dropping significantly, the network remains overall extremely confident in its predictions.}
    \label{fig:Shift-corrupt-mnist}
\end{figure*}

This last experimental subsection is dedicated to \emph{distribution shift}. Distribution shifts share similar ideas with OOD-detection, in the sense that the network will be confronted to samples that are not following the training distribution. 
However, the difference lies in the fact that these new observations are shifted instances of observations that would be sampled with respect to the training distribution. 
In particular, shifted instances still have an underlying label that one may hope to recover. 
Formally, given a training distribution $\XX$ and a parametric family of shifts $(s_\gamma)_\gamma$ with the convention that $s_0 = \mathrm{id}$, a shifted sample with level of shift $\gamma$ is a sample $s_\gamma(x_1), \dots, s_\gamma(x_N)$, where $x_1, \dots, x_N \sim \XX$ with underlying labels $y_1, \dots, y_N$. 
For instance, given an image, one can apply a \emph{corruption shift} of parameter $\gamma = n \in \N$ where $s_n(x)$ consists of randomly switching $n$ pixels of the image $x$ ($x_{ij} \mapsto 1 - x_{ij}$). 
See the top row of \cref{fig:Shift-corrupt-mnist} for an illustration. 

Ideally, one would hope a trained network $F$ to be robust to shifts, that is $\argmax(F(x)) = \argmax(F(s_\gamma(x)))$. However, since the map $x \mapsto s_\gamma(x)$ cannot be inverted in general, one cannot realistically expect robustness to hold for high levels of shift. Here, we illustrate how TU can be used as a way to monitor the presence of shifts that would lead to a dramatic diminution of the network accuracy in situations where the network confidence would be helpless.

For this, we train a network on \texttt{MNIST} and, following the methodology presented in \cref{sec:methodology}, store the corresponding average persistence diagrams for the $10$ classes appearing in the training set. 
We then expose a batch of $1000$ observations from the test set containing $100$ instances of each class (that have thus not been seen by the network during the training phase) to the corruption shifts with various shift levels $n \in \{0, 10, 20, 50, 100, 150, \dots, 500\}$. 
For each shift level, we evaluate the distributions of TUs and confidences attributed by the network to each sample, along with the accuracy of the network over the batch. As illustrated in the second row of \cref{fig:Shift-corrupt-mnist}, as the batch shifts, the TU increases and the accuracy drops. 
However, the network confidence remains very close to $1$, making this indicator unable to account for the shift. In practice, one can monitor a network by routinely evaluating the distribution of TUs of a new batch (\textit{e.g.}, daily recorded data). 
A sudden change in this 1D distribution is likely to reflect a shift in the distribution of observations that may itself lead to a drop in accuracy (or the apparition of OOD samples as illustrated in \cref{subsec:ood}).

We end this subsection by stressing that the empirical relation we observe between the TU and the network accuracy cannot be guaranteed without further assumption on the law $(\XX, \YY)$. 
It however occurs consistently in our experiments (see the appendix for complementary experiments involving different types of shift and network architectures). 
Studying the theoretical behavior of the TU and activation graphs in general will be the focus of further works.

\section{Conclusion and perspectives}

Monitoring trained neural networks deployed in practical applications is of major importance and is challenging when facing samples coming from a distribution that differs from the training one. 
While previous works focus on improving the behavior of the network confidence, in this article we propose to investigate the whole network instead of restricting to its final layer. 
By considering a network as a sequence of bipartite graphs on top of which we extract topological features, we introduce the Topological Uncertainty, a tool to compactly quantify if a new observation activates the network in the same way as training samples did. 
This notion can be adapted to a wide range of networks and is entirely independent from the way the network was trained. 
We illustrate numerically how it can be used to monitor networks and how it turns out to be a strong alternative to network confidence on these tasks. We will make our implementation publicly available.

We believe that this work will motivate further developments involving Topological Uncertainty, and more generally activation graphs, when it comes to understand and monitor neural networks. In particular, most techniques introduced in recent years to improve confidence-based descriptors may be declined to be used with Topological Uncertainty. These trails of research will be investigated in future work.

%\clearpage
\bibliographystyle{named}
\bibliography{biblio}

\appendix 

\section{Elements of Topological Data Analysis and theoretical considerations}
\label{sec:supmat:theory}

\subsection{Construction of persistence diagrams for activation graphs.}
\begin{figure*}
    \centering
    \includegraphics[width=0.9\textwidth]{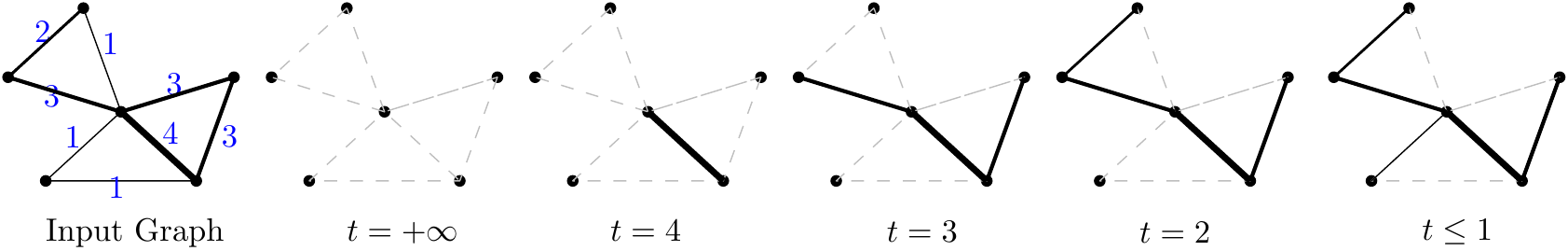}
    \caption{The super-level set filtration of a weighted graph, where we only draw edges that connect two connected components (not those creating loops). The filtration ends up with a Maximum Spanning Tree of the graph.}
    \label{fig:introTDA}
\end{figure*}

In this section, we provide a brief introduction to Topological Data Analysis (TDA) and refer to \cite{edelsbrunner2010computational,Oudot2015,chazal2017introduction} for a more complete description of this field. 

Let $X$ be a topological space and $f \colon X \to \R$ be a real-valued continuous function. The sublevel set of parameter $t$ of $(X,f)$ is the set $\FF_t := \{ x \in X \;:\; f(x) \leq t \} \subseteq X$. 
As $t$ increases from $-\infty$ to $+\infty$, one gets an increasing sequence (\textit{i.e.}, nested with respect to the inclusion) of topological spaces called the \emph{filtration} of $X$ by $f$. 
Given such a filtration, persistent homology tracks the times $t$ of appearance and disappearance of topological features (such as connected components, loops, cavities, etc.). For instance, a new connected component may appear at time $t_1$ in the sublevel set $\FF_{t_1}$, and may disappear---that is, it may get merged with an already present, other connected component---at time $t_2 \geq t_1$. 
We say that this connected component \emph{persists} over the interval $[t_1, t_2]$ and we store the pairs $(t_1, t_2)$ as a point cloud in the Euclidean plane called the \emph{persistence diagram} of the sublevel sets of $(X,f)$. Persistence diagrams of the \emph{superlevel sets} are defined similarly. We refer to \cref{fig:introTDA} for an illustration.

Now, given a graph $G = (V,E)$ and a weight function $w \colon E \to \R$, we build a superlevel set persistence diagram as follows: at $t=+\infty$ (\textit{i.e.},~at the very beginning), we add all the nodes $V$. 
Then, we build a sequence of subgraphs $G_t = \{ (V, E_t) \}$ with $E_t = \{e \in E\,:\, w(e) \geq t\}$, that is, we add edges in decreasing order with respect to their weights. In this process, no new connected component can be created (we never add new vertices), and each new edge inserted at time $t$ either connects two previously independent components leading to a point with coordinates $(+\infty, t)$ in the persistence diagram, or creates a new loop in the graph. 
In our construction, we propose to only keep track of connected components ($0$-dimensional topological features).
Since all points in the persistence diagram have $+\infty$ as first coordinate, we simply discard it and end up with a distribution of real numbers $t_1 \geq \dots \geq t_N$, where $N = |V|-1$, which exactly corresponds to the weights of a maximum spanning tree built on top of $G$, see for instance \cite[Proof of Lemma 2]{doraiswamy2020topomap} for a proof. 

Therefore, in our setting, we can compute the persistence diagram of an activation graph $G_\ell(x,F)$ connecting two layers, using the weight function $w \colon e \mapsto |W_\ell(i,j) x_\ell(i)|$, where $e$ is the edge connecting the $i$-th unit of the input layer to the $j$-th unit of the output layer.

\paragraph{A comparison with Neural persistence.} Given a bipartite graph, the construction we propose is exactly the one introduced in \cite{rieck2019neural}. The difference between the two approaches is that \textit{loc.~cit.}~works with the network weights $|W_\ell(i,j)|$ directly, obtaining one graph for each layer of the network $F$, while we use activation graphs $G_\ell(x,F)$ with weights $|W_\ell(i,j) x_\ell(i)|$, so that we obtain one graph \emph{per observation} $x$ for each layer of the network. Then, both approaches compute a persistence diagram (MST) on top of their respective graphs. The second difference is that, in order to make use of their persistence diagrams in a quantitative way, \cite{rieck2019neural} proposes to compute their \emph{total persistence}, which is exactly the distance between a given diagram and the empty one (forcing to match all the points of the given diagram onto the diagonal $\Delta$). Taking advantage of the fact that  our diagrams always have the same number of points (determined by the in- and out-layers sizes), we instead propose to compute distances to a reference diagram, namely the Fr\'echet mean of diagrams obtained on the training set. This allows us to compactly summarize information contained in the training set and use it to monitor the network behavior when facing new observations.

\subsection{Metrics and Stability results.}

\begin{figure}
    \centering
    \includegraphics[width=0.3\textwidth]{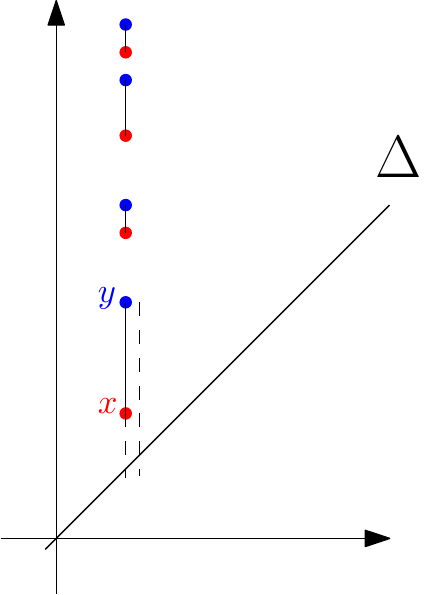}
    \caption{Two persistence diagrams whose points are distributed on the real line. Since we use $\| \cdot \|_1$ as a ground metric between diagram points, the distance between any two points $x$ and $y$ is always smaller than or equal to the sum of their distances to the diagonal. Therefore, an optimal partial matching between the two persistence diagrams can be obtained by simply matching points in increasing order (plain lines).}
    \label{fig:matching1D}
\end{figure}

\paragraph{Diagram metrics in our setting.} Let $\Delta = \{ (t,t) \in \R^2 \}$ denote the diagonal of the Euclidean plane. Let $\mu, \nu$ be two diagrams with points $X := \{x_1, \dots, x_n \} \subseteq \R^2$ and $Y := \{y_1, \dots, y_m\} \subseteq \R^2$, respectively\footnote{In general, two persistence diagrams $\mu$ and $\nu$ can have different numbers of points.}. 
Persistence diagrams are compared using optimal partial matching metrics. 
Namely, given a parameter $1 \leq p \leq +\infty$, the $p$-th diagram metric is defined by
\begin{align}
    \mathbf{d}_p(\mu,\nu) = \left(\inf_{\gamma} \sum_{x \in X \cup \Delta} \|x - \gamma(x) \|^p_1 \mathrm{d} x\right)^{1/p},
\end{align}
where $\gamma$ ranges over bijections between $X \cup \Delta$ and $Y \cup \Delta$. 
Intuitively, it means that one can either match points in $X$ to points in $Y$, or match points in $X$ (resp.\ in  $Y$) to their orthogonal projections onto the diagonal $\Delta$. 
Among all the possible bijections, the optimal one is selected and used for computing the distance.

This combinatorial problem is computationally expensive to solve (\textit{a fortiori}, so is the one of computing Fr\'echet means for this metric). 
Fortunately, in our setting, where diagrams are actually one-dimensional, that is, supported on the real line, and have the same number of points $N = d_{\ell} + d_{\ell+1} - 1$, computing the optimal bijection becomes trivial. 
Indeed, one can observe first that, since we use $\| \cdot \|_1$ as our ground metric on the plane, it is never optimal to match a point $x \in X$ (resp.~$y \in Y$) to the diagonal, since $\|x - y\|_1 \leq \|x - s(x)\|_1 + \|y-s(y)\|_1$, where $s$ denotes the orthogonal projection of a point onto the diagonal, see \cref{fig:matching1D}. 
Since the persistence diagrams (that have to be compared) always have the same number of points in our setting, we can simply ignore the diagonal. 
In this way, we end up with an optimal matching problem between 1D measures. 
In that case, it is well-known that the optimal matching is given by matching points in increasing order, see for instance \cite[\S 2.1]{santambrogio2015optimal}. 
Thus, the distance $\mathrm{Dist}$ introduced in the article is equal to the distance $\mathbf{d}_1$ up to a normalization term $1/N$. 
This normalization is meant to prevent large layers from artificially increasing the distance between diagrams (and thus the Topological Uncertainty) and helps making distances over layers more comparable. 
The same normalization appears in \cite{rieck2019neural}.

\paragraph{Stability.} Let $G = (V, E)$ be a graph, and $w, w' \colon E \to \R$ be two weight functions on its edges. In this context, the stability theorem (see for instance \cite{chazal2016structure}) in TDA states that
\begin{align}
    \mathbf{d}_\infty (\mathrm{Dgm}(G,w), \mathrm{Dgm}(G,w')) \leq \sup_{e \in E} | w(e) - w'(e) |.
    \label{eq:stabilityThm}
\end{align}
Now, let us consider a (sequential) neural network $F$ with $L$ layers and assume that all activations map $\sigma_\ell$ for $1 \leq \ell \leq L-1$ are $1$-Lipschitz\footnote{This is typically the case if one considers ReLU or sigmoid activation maps, and remains true if those are post-composed with $1$-Lipschitz transformations such as MaxPooling, etc.}. 
%Let $x, y \in \R^d$ and recall that $x_{\ell+1} = \sigma_\ell(W_\ell \cdot x_\ell + b_\ell)$ (resp.~$y_{\ell+1}$), with $x_1 = x$. 
We now introduce the notation 
     \[ A_\ell = \sup_{i_\ell, i_{\ell+1}} \sum_{i_\ell=1}^{d_\ell} |W_\ell(i_\ell, i_{\ell+1})|. \]  
Then, the map $x_\ell \mapsto \sigma_\ell(W_\ell x_\ell + b_\ell)$ is an $A_\ell$-Lipschitz transformation (for the $\| \cdot \|_\infty$ norm). Indeed, one has:
\begin{align}
    &\sup_{1 \leq i_{\ell+1} \leq d_{\ell+1}} \bigg\| \sigma_\ell\left( \sum_{i_\ell = 1}^{d_\ell} W_\ell(i_\ell,i_{\ell+1}) x_\ell(i_\ell) + b_\ell(i_\ell) \right) \\
    &- \sigma_\ell\left( \sum_{i_\ell = 1}^{d_\ell} W_\ell(i_\ell,i_{\ell+1}) y_\ell(i_\ell) + b_\ell(i_\ell) \right) \bigg\| \\
    \leq & \sup_{1 \leq i_{\ell+1} \leq d_{\ell+1}} \sum_{i_\ell = 1}^{d_\ell} |W_\ell(i_\ell,i_{\ell+1})| |x_\ell(i_\ell) - y_\ell(i_\ell) | \\
    \leq & A_\ell \| x_\ell - y_\ell\|_\infty.
\end{align}
Therefore, the map $x \mapsto x_\ell$ is Lipschitz with Lipschitz constant $\AA_\ell := \prod_{\ell' \leq \ell} A_{\ell'}$. 
Now, using the stability theorem, we know that for $x,y$ two observations, one has
\begin{align}
    \mathbf{d}_\infty (D_\ell(x,F), D_\ell(y,F)) \leq \AA_\ell \|x - y\|_\infty.
\end{align}
Finally, observe that $\frac{1}{N} \mathbf{d}_1 \leq \mathbf{d}_\infty$, so that the same result holds using the metric $\mathrm{Dist}$ introduced in the main paper. 

We end this subsection by highlighting that, contrary to most applications of TDA, having a low Lipschitz constant in our stability result is not crucial as we actually want to be able to separate observations based on their Topological Uncertainties. 
Nevertheless, studying and proving results in this vein is of great importance in order to understand the theoretical properties of Topological Uncertainty, a project that we left for further work.

\section{Complementary experimental details and results}
\label{sec:supmat:expe}

\paragraph{Implementation.} 

Our implementation relies on \texttt{tensorflow 2} \cite{abadi2016tensorflow} for neural network training and on \texttt{Gudhi} \cite{maria2014gudhi} for persistence diagrams (MST) computation. 
Note that instantiating the \texttt{SimplexTree} (representation of the activation of the graph in \texttt{Gudhi}) from the matrix $(|W_\ell(i,j) x_\ell(i)|)_{ij}$ is the limiting factor in terms of running time in our current implementation, taking up to $0.9$ second on a $784 \times 512$ dense matrix for instance\footnote{On a \texttt{Intel(R) Core(TM) i5-8350U CPU 1.70GHz}, average over 100 runs.}. 
Using a more suited framework\footnote{In our experiments, \texttt{Gudhi} appeared however to be faster than \texttt{networkx} and \texttt{scipy}, two other standard libraries that allow graph manipulation.} to turn a matrix into a MST may drastically improve the computational efficiency of our approach. 
As a comparison, extracting a MST of a graph using \texttt{Gudhi} only takes about $0.03$ second.

\subsection{Datasets.}

\paragraph{Datasets description.} 

The \texttt{MNIST} and \texttt{FMNIST} datasets \cite{mnist,fmnist} both consist of training data and $10,000$ test data that are grey-scaled images of shape $28 \times 28$), separated in $10$ classes, representing hand-written digits and fashion articles, respectively. 

The \texttt{CIFAR10} dataset \cite{cifar10} is a set of RGB images of shape $32 \times 32 \times 3$, separated in $10$ classes, representing natural images (airplanes, horses, etc.).

The \texttt{SVHN} dataset (Street View House Number) \cite{svhn} is a dataset of RGB images of shape $32 \times 32 \times 3$, separated in $10$ classes, representing digits taken from natural images.

The \texttt{DTD} dataset (Describable Texture Dataset) \cite{dtd} is a set of RGB images whose shape can vary between $300 \times 300 \times 3$ and $640 \times 640 \times 3$, separated in 47 classes. 

Finally, \texttt{Gaussian} and \texttt{Uniform} refers to synthetic dataset of images of shape $32 \times 32 \times 3$ where pixels are sampled \textit{i.i.d} following a normal distribution $\mathcal{N}(0.5, 0.1)$ and a uniform distribution on $[0,1]$, respectively.
\medbreak
The \texttt{MUTAG} dataset is a small set of graphs (188 graphs) coming from chemestry, separated in $2$ classes. 
Graphs have on average $17.9$ vertices and and $19.8$ edges.
They do not contain nodes or edges attributes and are undirected.

The \texttt{COX2} dataset is a small set of graphs (467 graphs) coming from chemestry, separated in $2$ classes. 
Graphs have on average $41.2$ vertices and and $43.7$ edges. They do not contain nodes or edges attributes and are undirected.

\paragraph{Preprocessing.} 

All natural images datasets are rescaled to have  coordinates in $[0,1]$ (from $[0,255]$). The coordinates of images belonging to the \texttt{Gaussian} dataset are thresholded to $[0,1]$.
When feeding a network trained on \texttt{CIFAR10} with smaller grey-scaled images (from \texttt{MNIST} and \texttt{FMNIST}), we pad the smaller images with zeros (by adding two columns (resp.~rows) on the left and on the right of the image) and repeat the grey-scale matrix along the three channels. 
Conversely, we randomly sub-sample $32 \times 32 \times 3$ patches for images coming from the \texttt{DTD} dataset, as typically done in similar experiments.
\medbreak
The preprocessing of graphs datasets is slightly more subtle. 
In order to describe graphs as Euclidean vectors, we follow a procedure described in \cite{carriere2020perslay}. 
We compute the first $30$ eigenvalues of their normalized Laplacian (or pad with $0$s if there is less than $30$ eigenvalues). 
Additionally, we compute the quantile of the \emph{Heat Kernel Signature} of the graphs with parameter $t=10$ as suggested in \cite{carriere2020perslay}. 
This processing was observed to be sufficient to reach fairly good accuracies on both \texttt{MUTAG} and \texttt{COX2}, namely $\sim 88$\% on \texttt{MUTAG} and $\sim 78\%$ on \texttt{COX2}, about $2-3\%$ below the state-of-the-art results reported in \cite[Table 2]{carriere2020perslay}.

\subsection{Networks architectures and training.}

Depending on the type of data and the dataset difficulty, we tried our approach on different architectures. 

\begin{itemize}
    \item For \texttt{MUTAG} and \texttt{COX2}, we used a simple network with one-hidden layer with 32 units and ReLU activation. The input and the hidden layer are followed by Batch normalization operations. 
    \item For networks trained on \texttt{MNIST} and \texttt{FMNIST}, we used the reference architecture described at \url{https://github.com/pytorch/examples/tree/master/mnist} (although reproduced in \texttt{tensorflow 2}). 
    When computing TU on top of this architecture, we only take the two final fully-connected layers into account (see the paragraph {\bf Remarks} in the main paper).
    \item For networks trained on \texttt{CIFAR-10}, we used the architecture in \url{https://www.tensorflow.org/tutorials/images/cnn}. 
\end{itemize}
All networks are trained using the ADAM optimizer \cite{kingma2014adam} (from \texttt{tensorflow}) with its defaults parameters, except for graph datasets where the learning rate is taken to be $0.01$. 

\subsection{Complementary experimental results.}

We provide in this subsection some complementary experimental results and discussions that aim at putting some perspective on our work. 

\paragraph{About trained network selection.}

First, we stress that the results we obtain using either the TU or the confidence can probably be improved by a large margin\footnote{Our scores are however in the same range as those reported by \cite{ramamurthy2019topological}, namely $+10.5\%$ on \texttt{MNIST} and $+23\%$ on \texttt{Fashion-MNIST} in their experimental setting.}: if the method was perfectly reliable, one would expect a $\sim +50\%$ increase between the accuracy obtained by a network with low TU (resp.~high confidence) and the accuracy obtained by a network with high TU (resp.~low confidence). 
Indeed, this would indicate that the score clearly separates networks that perfectly split the input dataset (leading to an accuracy of $\sim 100\%$) and those which classify all the points in a similar way (accuracy of $50\%$). 
%Theo: should we give more details?...

\paragraph{OOD detection: complementary illustrations.}

\begin{figure}
    \centering
    \includegraphics[width=\columnwidth]{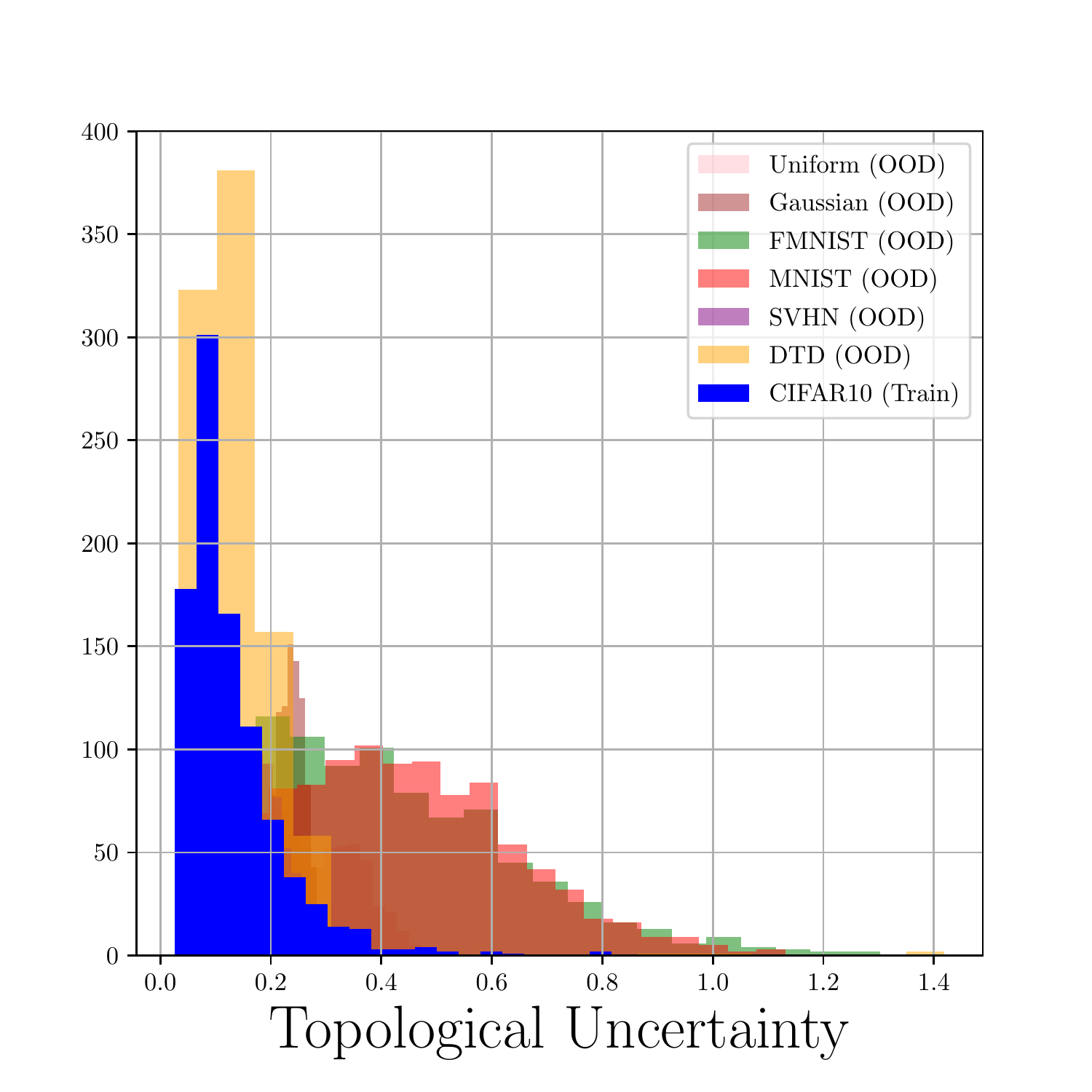}
    \caption{Distribution of TU for the \texttt{CIFAR-10} dataset (training set) and OOD datasets.}
    \label{fig:cifar10distribTU}
\end{figure}

\begin{figure}
    \centering
    \includegraphics[width=\columnwidth]{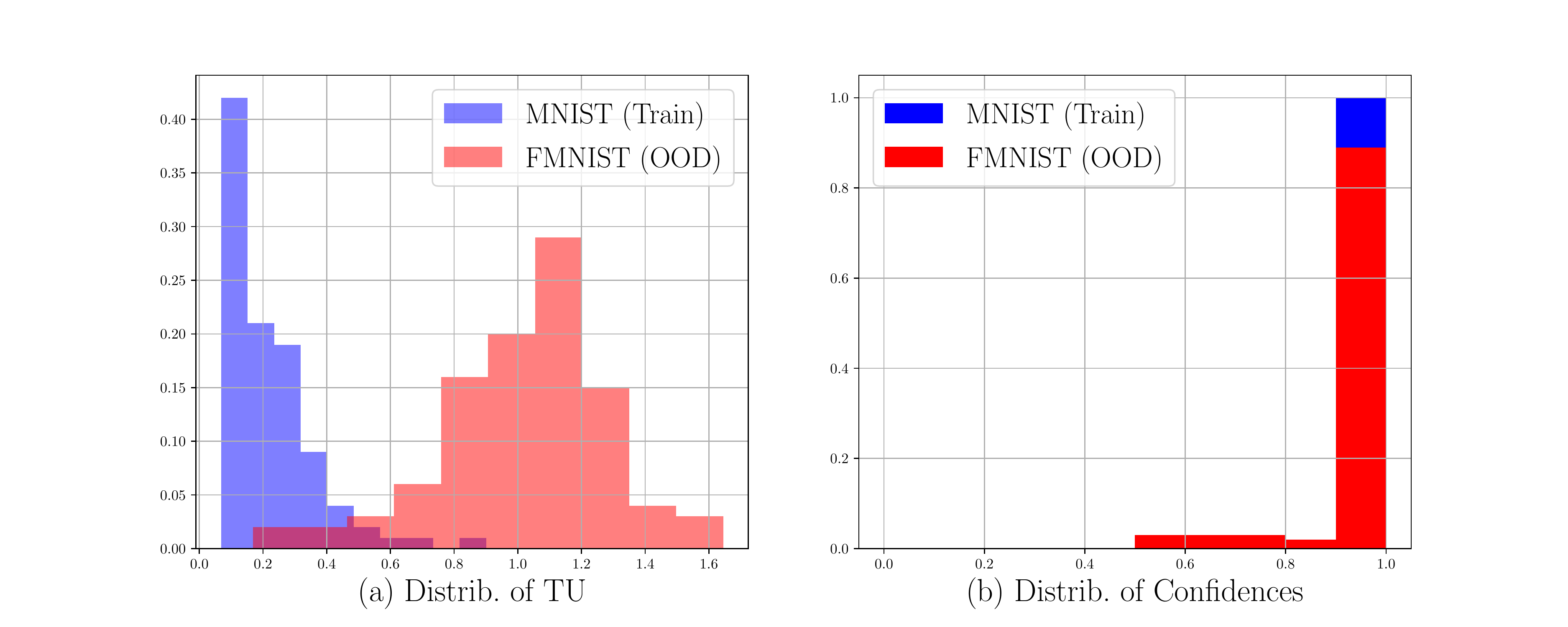}
    \caption{Distributions of TU and Confidence for a network trained on the \texttt{MNIST} dataset, using \texttt{FMNIST} as an OOD dataset.}
    \label{fig:mnistood}
\end{figure}

\cref{fig:cifar10distribTU} shows the distribution of the TU for a network trained on the \texttt{CIFAR-10} datasets (from which we store average persistence diagrams used to compute the TU) and a series of OOD datasets. 
Although the results are less conclusive that those obtained on graph datasets, the distribution of TUs on training data remains well concentrated around low values while being more spread for OOD datasets (with the exception of the \texttt{DTD} dataset, on which our approach indeed performs poorly), still allowing for a decent detection rate. 
\cref{fig:mnistood} provides a similar illustration for a network trained on \texttt{MNIST} using \texttt{FMNIST} as an OOD dataset. On this example, interestingly, the confidence baseline \cite{hendrycks2017baseline} achieves a perfect FPR at 95\% TPR of $0\%$ and an AUC of $64$, while our TU baseline achieves a FPR at 95\% of $5\%$ (slightly worse than the confidence baseline) but an overall good AUC of $98$.

\paragraph{Sensitivity to shift under Gaussian blur.} 

\begin{figure*}[th]
    \centering    
    \includegraphics[width=0.9\textwidth]{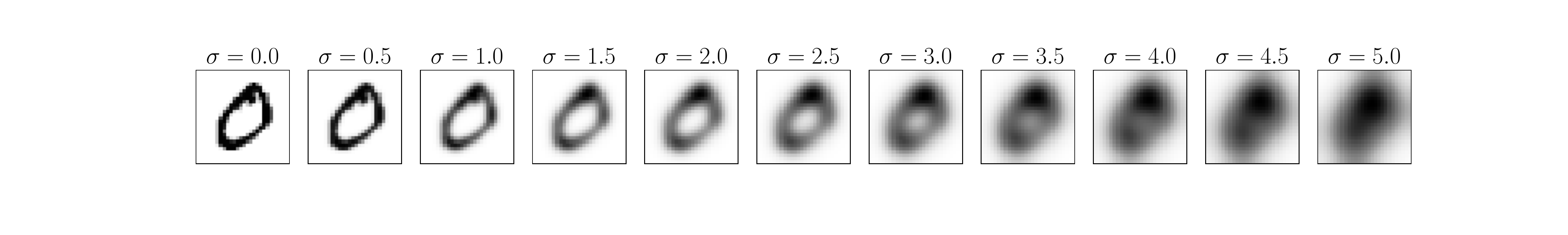}
    \includegraphics[width=0.8\textwidth]{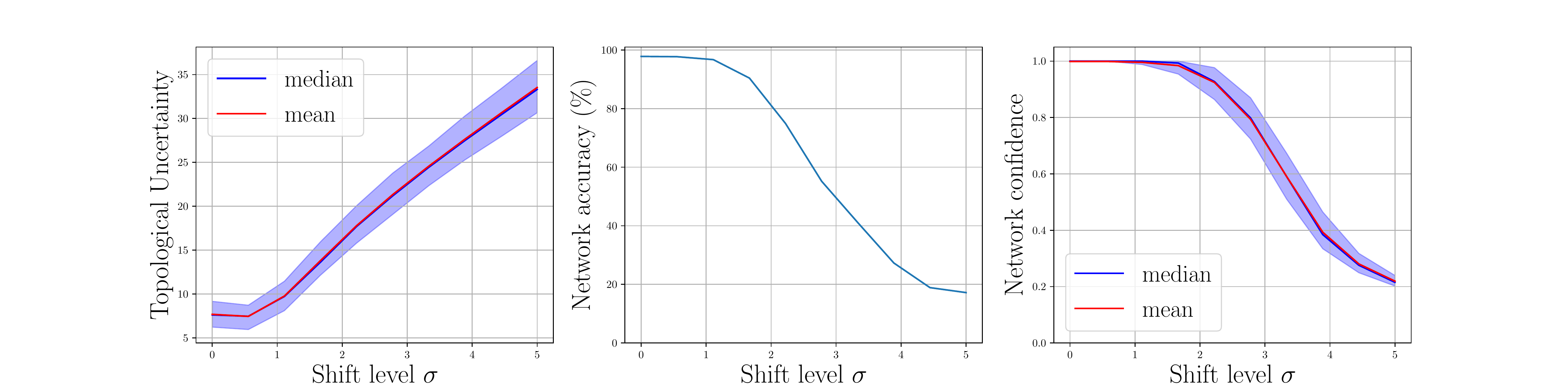}
    \vspace{-0.3cm}
    \caption{\textit{(Top row)} A $0$ digit from the \texttt{MNIST} dataset exposed to increasing level of shift (Gaussian blur). \textit{(Bottom row)}, \textit{(Left)}. The TU (with $0.1$ and $0.9$ quantiles) of corrupted inputs in the \texttt{MNIST} dataset with respect to the corruption level $\sigma$. \textit{(middle)} The accuracy of the network on these data (that is, the proportion of observations that are still correctly classified). \textit{(right)} The confidence of the network in its predictions.}
    \label{fig:supmat:gaussian_blur}
\end{figure*}

\cref{fig:supmat:gaussian_blur} illustrates the behavior of TU, accuracy, and network confidence when data from the \texttt{MNIST} are exposed to Gaussian blur of variance $\sigma \in [0,5]$ (see the top row for an illustration). 
The increase in TU accounts for the shift in the distribution in a convincing way. 
Interestingly, on this example, the confidence reflects the shift (and the drop in accuracy) in a satisfactory way, contrary to what happens when using a corruption shift. 
\end{document}